\documentclass[12pt]{article}
\newtheorem{remark}{Remark}[section]

\usepackage{longtable}
\usepackage{wrapfig}
\usepackage{lscape}
\usepackage{lipsum}
\usepackage{setspace}
\usepackage{setspace}
\usepackage{rotating}
\usepackage{epstopdf}
\usepackage{graphicx}
\usepackage[font=scriptsize]{caption}
\usepackage{color}
\usepackage{amssymb}
\usepackage{sansmath}
\usepackage{amsmath}
\usepackage{array}
\usepackage{multirow}
\usepackage{epsfig}
\usepackage{latexsym}
\usepackage[T1]{fontenc}
\usepackage[utf8]{inputenc}
\usepackage{authblk}
\usepackage{mathtools}
\usepackage[autostyle]{csquotes}
\date{}
\title{Robust mixture modelling using \\
sub-Gaussian $\alpha$-stable distribution}
\author[]{Mahdi Teimouri}
\author[]{Saeid Rezakhah}
\author[]{Adel Mohammdpour}
\affil[]{\small{Department of Statistics, Faculty of Mathematics and Computer Science, \\Amirkabir University of Technology (Tehran Polytechnic), 424 Hafez Ave., Tehran 15914, Iran}}
\begin{document}
\maketitle{}
\noindent{}{\bf{Abstract:}} Heavy-tailed distributions are widely used in robust mixture modelling due to possessing thick tails. As a computationally tractable subclass of the stable distributions, sub-Gaussian $\alpha$-stable distribution received much interest in the literature. Here, we introduce a type of expectation maximization algorithm that estimates parameters of a mixture of sub-Gaussian $\alpha$-stable distributions. A comparative study, in the presence of some well-known mixture models, is performed to show the robustness and performance of the mixture of sub-Gaussian $\alpha$-stable distributions for modelling, simulated, synthetic, and real data.\\\\
\noindent{}{\bf{Keyword:}}~Clustering; EM algorithm; Monte Carlo; Mixture models; Robustness; Stable distribution; sub-Gaussian $\alpha$-stable distribution
\\\\
\section{Introduction}
\setcounter{equation}{0}
Finite mixture models are in fact a convex combination of two or more probability density functions. As the most critical application, these models received much interest in the model-based clustering which focuses mainly on the mixture of Gaussian distributions. Despite the popularity of Gaussian-based clustering, this algorithm shows poor performance in the presence of outliers. So, robust mixture models are becoming increasingly popular to overcome this issue. Some of these models aim to tackle tail-weight, \cite{Andrews2012}, \cite{Browne2015}, \cite{Peel2000}, and \cite{Zhang2010}; some deal with skewness, \cite{Basso2010}, \cite{Cabral2012}, \cite{Lee2013}, and \cite{Vrbik2012}.
Stable distributions have received extensive use in vast majority of fields such as finance, and telecommunications, \cite{Mittnik2003}, \cite{Nikias1995}, \cite{Nolan2003}, \cite{Ortobelli2010}, and \cite{Rachev2000}. Statistical modelling of datasets gathered from these fields using normal distribution is quite improper because of heavy tails. Except for three cases, probability density function (pdf) of the class of stable distributions has not closed form. As a computationally tractable subclass of the multivariate stable distribution, the sub-Gaussian $\alpha$-stable distribution can account for modelling processes with outliers. The sub-Gaussian $\alpha$-stable distributions have received much interest in finance and portfolio optimization, \cite{Mittnik2003}, \cite{Nolan2003}, and \cite{Omelchenko2010}. So, several attempts have been made to estimate the parameters of sub-Gaussian $\alpha$-stable distribution. Among them we cite \cite{Bodnar2011}, \cite{Nolan2013}, \cite{Omelchenko2014}, and \cite{Ravishankera1999}. The sub-Gaussian $\alpha$-stable distribution allows for heavier tails than Student's $t$ distribution; it can be used as a more flexible tool robust model-based clustering. On the other hand, other approaches that have been developed for estimating the parameters of sub-Gaussian $\alpha$-stable distribution have no possibility of being extended for the mixture of sub-Gaussian $\alpha$-stable distributions. This motivated us to develop a method to estimating the parameters of the mixture of sub-Gaussian $\alpha$-stable distributions. It should be noted that idea of the Bayesian approach for estimating the parameters of mixture of sub-Gaussian $\alpha$-stable distribution suggested in \cite{Gonzalez2010} and they did not follow it. Our investigations reveal that the proposed EM algorithm shows better performance, regarding execution time than the Bayesian paradigm.
The structure of this note is as follows. In what follows, we give some preliminaries. The Proposed EM algorithm is described in Section 3. Section 4 is devoted to performance analysis of the proposed EM algorithm through simulation, real data, and synthetic data.
\section{Preliminaries}
Let $\boldsymbol{Y}_i=(Y_{i1},\dots,Y_{id})^T$ be a sub-Gaussian $\alpha$-stable random vector. Then a random sample
\begin{equation}\label{def2}
\boldsymbol{Y}_i\mathop=\limits^L \boldsymbol{\mu}+\sqrt{P}_i\boldsymbol{G}_i,~~i=1,\dots,n,
\end{equation}
where $\boldsymbol{G}_i=(G_{i1},\dots,G_{id})^T$ is a zero-mean Gaussian random vector with a positive definite symmetric $d \times d$ shape matrix $\Sigma$, $P_i$ follows $S\bigl(\alpha/2,\left(\cos(\pi\alpha/4)\right)^{2/\alpha},1,0\bigr)$, $\boldsymbol{\mu} \in \mathbb{R•}^d$ is a location parameter, and $0<\alpha< 2$. Here, $?\mathop=\limits^L?$ denotes equality in distribution and $P_i$ and $\boldsymbol{G}_i$ are statistically independent. The corresponding observed values of $\boldsymbol{Y}_i$, $\boldsymbol{G}_i$, and ${P}_i$ are $\boldsymbol{y}_i=(y_{i1},\dots,y_{id})^T$, $\boldsymbol{g}_i=(g_{i1},\dots,g_{id})^T$, and ${p}_i$, respectively, for $i=1,\dots,n$, \cite{Samorodnitsky1994}.
If $f(\boldsymbol{y}_i|\alpha_j, \boldsymbol{\Sigma}_j, \boldsymbol{\mu}_j)$ denotes the pdf of $\boldsymbol{Y}_i$ with parameters $\alpha_j$, $\boldsymbol{\Sigma}_j$, and $\boldsymbol{\mu}_j$ at point $\boldsymbol{y}_i$, then the pdf of a $K$-component sub-Gaussian $\alpha$-stable mixture model, $g(\boldsymbol{y}_i|\theta)$, has the form
\begin{equation}\label{mix}
g(\boldsymbol{y}_i|\theta)=\sum_{j=1}^{K}w_j f(\boldsymbol{y}_i|\alpha_j, \boldsymbol{\Sigma}_j, \boldsymbol{\mu}_j),~~i=1,\dots,n,
\end{equation}
where $\theta=(\alpha_j, \Sigma_j, \mu_j; j=1,\dots,K)$, $n$ is the sample size, and $w_j$s are non-negative mixing parameters that sum to one, i.e. $\sum_{j=1}^{K} w_j=1$. Hereafter, we use notation SG$\boldsymbol{\alpha}$SM as a symbol for $K$-component sub-Gaussian $\alpha$-stable mixture distribution in which $\boldsymbol{\alpha}=(\alpha_1,\dots,\alpha_K)^T$. Identifiability of the SG$\boldsymbol{\alpha}$SM distribution with pdf in (\ref{mix}) is valid from \cite{Holzmann2006}.
\par
Missing or incomplete observations frequently occur in the statistical studies. The EM algorithm, introduced in \cite{Dempster1977}, is a popular inferential tool for such a situation. The application of EM technique also includes the cases that we encounter the latent variables or models with random parameter provided that they are formulated as a missing value problem. Let $L_c(\theta)=f(\boldsymbol{y},\boldsymbol{z}|\theta)$ be the complete data likelihood function in which $\boldsymbol{y}$ and $\boldsymbol{z}$ denote the vector of observed and latent observations, respectively. The EM algorithm works iteratively by maximizing the conditional expectation, $Q\left(\theta |\theta^{(t)}\right)$, of the complete log-likelihood function given the available data and a current estimate $\theta^{(t)}$ of the parameter.
Each iteration of EM algorithm involves two steps as
the E-step (computing $Q\left(\theta|\theta^{(t)}\right)$ at $t$-th iteration) and the M-step (maximizing $Q\left(\theta |\theta^{(t)}\right)$ with respect to $\theta$ to get $\theta^{(t+1)}$). The E- and M-steps are repeated until convergence occurs.
\par As the M-step of EM algorithm is analytically intractable, we imply this step with a sequence of conditional maximization, known as CM-step. This procedure is known as ECM algorithm, \cite{Meng1993}. A faster extension of EM algorithm, i. e. the ECME algorithm introduced in \cite{Liu1994}. It should be noted that all the EM, ECM, and ECME have the same E-step. The ECME algorithm works by maximizing the constrained $Q\left(\theta |\theta^{(t)}\right)$ via some CM-steps and maximizing the constrained marginal likelihood function and some constraints on the parameters, \cite{Basso2010}.
In cases where implementation of the EM algorithm is difficult, another extension of this algorithm, called stochastic EM (SEM) is useful, \cite{Celeux1985}. We imply SEM by simulating missing data from conditional density of $P^{(t)}_i$ given $y_i$ and $\theta^{(t)}$ with pdf $f(p^{(t)}_i|y_i, \theta^{(t)})$; for $i=1,\dots,n$, and substituting its sample $\boldsymbol{P}=(P^{(t)}_{1},\dots,P^{(t)}_{n})^T$ into the complete likelihood function. Then, we apply EM algorithm for the pseudo-complete sample $P^{(t)}_{1},\dots,P^{(t)}_{n}$. This process is repeated until convergence occurs for the distribution of the $\{\theta^{(t+1)}\}$. Under some mild regularity conditions, $\{\theta^{(t+1)}\}$ constitutes a Markov chain that converges to a stationary distribution, \cite{Ip1994}. The SEM is generally very robust to the starting values, and the number of iterations is determined via an exploratory data analysis approach such as, graphical display, \cite{Ip1994}. Using a burn-in of $N_0$ iterations, the sequence $\{\theta^{(t)}\}$ is expected to be close to some stationary point. After a sufficiently large number of iterations, say $N$, the SEM estimation of $\theta$ is given by
\begin{equation*}
\hat{\theta}=\frac{1}{N-N_0}\sum_{t=N_0+1}^{N} \theta^{(t)},
\end{equation*}
where $N_0$ is burn-in size. Upon above statements, each iteration of the SEM algorithm consists of two steps as follows.
\begin {enumerate}
\item Stochastic imputation (S-) step: Substitute the simulated missing values in the pseudo-complete log-likelihood function at $t$-th iteration.
\item Maximization (M-) step: Find a $\theta$, say $\theta^{(t+1)}$, which maximizes pseudo-complete log-likelihood function at $t$-th iteration.
\end {enumerate}
The S- and M-steps, in above, are repeated until convergence occurs.
\section{EM algorithm for SG$\alpha$SM}
We consider $\boldsymbol{y}_1,\dots,\boldsymbol{y}_n,\boldsymbol{z}_1,\dots,\boldsymbol{z}_n,{p}_1,\dots,{p}_{n}$ as the complete data corresponding to (\ref{mix}) where $\boldsymbol{y}_1,\dots,\boldsymbol{y}_n$ are observed data, $\boldsymbol{z}_1,\dots,\boldsymbol{z}_n$, are component labels and ${p}_1,\dots,{p}_{n}$ are missing observations. That is, if the $j$-th component, for $j=1,\dots,K$, of $\boldsymbol{Z}_{i}=(Z_{i1},\dots, Z_{iK})^T$ is one, then the other components are zero and $i$-th observation is coming from $j$-th component. This occurs with probability $w_j$. We have
\begin{equation} \label{pcon0}
\boldsymbol{Y}_i|P_i=p_i,Z_{ij}=1\sim N\left(\mu_{j}, p_i \Sigma_{j} \right),
\end{equation}
independently and
\begin{equation}\label{pcon}
P_i | Z_{ij}=1\sim S\Bigl(\alpha_{j}/2,1,\bigl(\cos(\pi \alpha_{j}/4)\bigr)^{2/\alpha_{j}},0\Bigr),
\end{equation}
for $j=1,\dots, K$ and $i=1,\dots,n$. It should be noted that given ${Z}_{ij}=1$, $P_i$s are independent.
So, the complete data density function can be represented as
\begin{equation}
f_c(\boldsymbol{y}_1,\dots,\boldsymbol{y}_n,\boldsymbol{z}_1,\dots,\boldsymbol{z}_n,{p}_1,\dots,{p}_{n}|\theta)=\prod^{n}_{i=1} f(\boldsymbol{y}_{i}, p_i, \boldsymbol{z}_{i}|\theta),
\end{equation}
where
\begin{align}
f(\boldsymbol{y}_{i}, p_i, \boldsymbol{z}_{i}|\theta)=&f_{\boldsymbol{Z}_{i}}(\boldsymbol{z}_{i})
f_{P_{i}|\boldsymbol{Z}_{i}}(p_i|\boldsymbol{z}_{i})
f_{\boldsymbol{Y}_{i}|P_{i},\boldsymbol{Z}_{i}}(\boldsymbol{y}_{i}|p_{i},\boldsymbol{z}_{i})\nonumber\\
=&w_{1}^{z_{i1}}\times \dots \times w_{K}^{z_{iK}}
\times\left\{f_{P_{i}|Z_{i1}}(p_i|{z}_{i1},\alpha_{1})\right\}^{z_{i1}}\times\dots
\times\left\{f_{P_{i}|Z_{iK}}(p_i|{z}_{iK},\alpha_{K})\right\}^{z_{iK}}\nonumber\\
&\times \left\{f_{\boldsymbol{Y}_{i}|P_{i},{Z}_{i1}}(\boldsymbol{y}_{i}|p_i,{z}_{i1},\boldsymbol{\mu}_{1}, \Sigma_1)\right\}^{z_{i1}}\times \dots \times \left\{f_{\boldsymbol{Y}_{i}|P_{i},{Z}_{i1}}(\boldsymbol{y}_{i}|p_i,{z}_{iK},\boldsymbol{\mu}_{K}, \Sigma_K)\right\}^{z_{iK}}\nonumber\\
=&\prod^{K}_{j=1}\left\{w_{j}f_{P_{i}|{Z}_{ij}}(p_i|{z}_{ij},\alpha_j)f_{\boldsymbol{Y}_{i}|P_{i},{Z}_{ij}}(\boldsymbol{y}_{i}|p_i, {z}_{ij},\boldsymbol{\mu}_{j}, \Sigma_j)\right\}^{z_{ij}}\nonumber,
\end{align}
where $z_{ij} \in \{0,1\}$ and $\sum_{j=1}^{K}z_{ij}=1$. It follows, from relations (\ref{pcon0}) and (\ref{pcon}), that the complete data log-likelihood $l_{c}(\theta)$ has the following representation
\begin{align}\label{lcom}
l_{c}(\theta)={C}&+\sum_{j=1}^{K} \sum_{i=1}^{n} z_{ij}\log w_j+\sum_{j=1}^{K}\sum_{i=1}^{n} z_{ij} \log f_{P}(p_i|\alpha_j)\nonumber\\
&-\frac{n}{2}\sum_{j=1}^{K}\sum_{i=1}^{n} z_{ij} \log |\Sigma_j| -\frac{1} {2}\sum_{j=1}^{K}\sum_{i=1}^{n}z_{ij}\frac{(\boldsymbol{y}_i-\boldsymbol{\mu}_j)^T \Sigma^{-1}_j (\boldsymbol{y}_i-\boldsymbol{\mu}_j)}{p_{i}},
\end{align}
where ${C}$ is a constant independent of $\theta=(\alpha_j, \Sigma_j, \mu_j; j=1,\dots,K)$. Considering $l_{c}(\theta)$ as a function of component label and missing variable $P_i$, its conditional expectation $Q\left(\theta|\theta^{(t)}\right)=E_P\left(l_c(\theta)|\boldsymbol{y}, \theta^{(t)}\right)$ becomes
\begin{align*}
Q\left(\theta \Big|\theta^{(t)}\right)=\text{C}&+\sum_{j=1}^{K} \sum_{i=1}^{n} e^{(t)}_{1ij}\log w_j+\sum_{j=1}^{K}\sum_{i=1}^{n} e^{(t)}_{1ij}E \left[\log f_P(p_i|\alpha_j)\Big | \boldsymbol{y}_i,\theta^{(t)}\right]\nonumber\\
&-\frac{n}{2}\sum_{j=1}^{K} \log |\Sigma_j|-\frac{1}{2}\sum_{j=1}^{K}\sum_{i=1}^{n}e^{(t)}_{1ij}e^{(t)}_{2ij} (\boldsymbol{y}_i-\boldsymbol{\mu}_j)^T \Sigma^{-1}_j (\boldsymbol{y}_i-\boldsymbol{\mu}_j),
\end{align*}
where
\begin{align}\label{e1ij}
e^{(t)}_{1ij}=E\left(Z_{ij}|\boldsymbol{y}_i, \boldsymbol{\mu}^{(t)}_j, \Sigma^{(t)}_j, \alpha^{(t)}_j\right)=\frac{w^{(t)}_j f({\boldsymbol{y}}_{i};\alpha^{(t)}_j, \Sigma^{(t)}_j, \boldsymbol{\mu}^{(t)}_{j})}{\sum_{j=1}^{K} f({\boldsymbol{y}}_{i};\alpha^{(t)}_j, \Sigma^{(t)}_j, \boldsymbol{\mu}^{(t)}_{j})},
\end{align}
in which $f({\boldsymbol{y}}_{i};\alpha^{(t)}_j, \Sigma^{(t)}_j, \boldsymbol{\mu}^{(t)}_{j})$ is pdf of a sub-Gaussian $\alpha$-stable random vector $\boldsymbol{Y}_i$ defined in (\ref{def2}) and
\begin{align}\label{e2ij}
e^{(t)}_{2ij}=E\left(P_{i}^{-1}|\boldsymbol{y}_i, \boldsymbol{\mu}^{(t)}_j, \Sigma^{(t)}_j, \alpha^{(t)}_j\right).
\end{align}
{{\bf{E-step:}}} The E-step is complete by computing $e^{(t)}_{1ij}$ and $e^{(t)}_{2ij}$. Details for computing these quantities are given
in Appendix 1. For this, we use package $\mathsf{STABLE}$, \cite{stable}.\\
{{\bf{M-step:}}} The M-step of the EM algorithm updates the weight and location parameters of $j$-th component in $(t+1)$-th iteration as follows.
\begin{equation}\label{what}
{w}^{(t+1)}_j=\frac{1}{n}\sum^{n}_{i=1}e^{(t)}_{1ij},
\end{equation}
\begin{equation}\label{muhat}
{\boldsymbol{\mu}}^{(t+1)}_j=\frac{\sum_{i=1}^{n} e^{(t)}_{1ij} e^{(t)}_{2ij} \boldsymbol{y}_{i}}{\sum_{i=1}^{n} e^{(t)}_{1ij}e^{(t)}_{2ij}},
\end{equation}
The shape matrix can be updated analytically in M-step, but we prefer to update it along with the tail index in a CM-step (this reduces computational costs). At $(t+1)$-th iteration, the updated quantities $e_{1ij}^{(t+1)}$, $e_{2ij}^{(t+1)}$, $w_{j}^{(t+1)}$, and $\mu_{j}^{(t+1)}$; for $j=1,\dots, K$ and $i=1,\dots, n$ are evaluated from (\ref{e1ij})-(\ref{muhat}), respectively. Using these updates, we follow to update ${\alpha}_j$ and ${\Sigma}_j$ as $\alpha^{(t)}_j$ and $\Sigma_{j}^{(t)}$; for $j=1,\dots, K$ in the CM-step. It should be noted that the CM-step can be implemented ?using numerical optimization tools. But the use of the Remark \ref{remark}, which suggests to use a stochastic EM (SEM) algorithm, leads to a mathematically and computationally tractable CM-step.
\begin{remark}\label{remark}
Suppose that $P$ is a positive stable random variable with tail index $\alpha/2$ and $E$ is an exponential random variable with mean one. Then, the quotient $E/P$ follows a Weibull distribution, independent of $E$, with shape parameter $\alpha/2$ and scale parameter unity, \cite{Samorodnitsky1994}.
\end{remark}
\begin{itemize}
\item {{\bf{First step of CM:}}} We consider $K$ groups $I_1,\dots, I_K$. Let $e^{(t+1)}_{1i}=(e^{(t+1)}_{1ij},\dots, e^{(t+1)}_{1iK})$, where $e^{(t)}_{1ij}$ is defined by (\ref{e1ij}). If the $j$-th component of $e^{(t+1)}_{i1}$ is larger than the other components, then $\boldsymbol{Y}_i$ is assigned to the $j$-th group $I_j$; for $i=1,\dots, n$, $j=1,\dots, K$. Now, $I_j$ whose size is $n_j$ consists of $\boldsymbol{Y}^{j}=(\boldsymbol{Y}^{j}_{1},\dots,\boldsymbol{Y}^{j}_{n_{j}})$ and $\sum_{j=1}^{K}n_j=n$. Using (\ref{def2}) and remark \ref{remark}, it follows that
\begin{align*}
\boldsymbol{\cal{Y}}^{j}_{i}=
\frac{\boldsymbol{Y}^{j}_{i}-\boldsymbol{\mu}^{(t+1)}_j}{\sqrt{E_i}}\mathop=\limits^L
\frac{\sqrt{P_i}N_i\left(0,\Sigma^{(t+1)}_j\right)}{\sqrt{E_i}},~~i=1,\dots,n_j,
\end{align*}
where $E_i$ is an exponential random variable with mean one independent of $P_i$, and $N_i$ is a $d$-dimensional zero-mean normal random vector with shape matrix $\Sigma^{(t+1)}_j$. It is easy to check that,
\begin{align*}
\frac{\sqrt{P_i}}{\sqrt{E_i}} \mathop=\limits^L V^j_i,
\end{align*}
where $V^j_i$ follows a Weibull distribution with shape parameter $\alpha_j/2$ and scale parameter one. Therefore,
\begin{align*}
\boldsymbol{\cal{Y}}^{j}_{i}|V^{j}_i=v^j_{i} &\sim N\left(0, \frac{\Sigma^{(t+1)}_j}{(v^j_{i})^2}\right),\nonumber\\
V^{j}_i &\sim \text{Weibull}(\alpha_j , \text{scale}=1),
\end{align*}
for $m=1,\dots, n_j$, $j=1,\dots, K$. By considering $v^{j}_i$ as the missing observation, the complete data log-likelihood of $j$-th group is
\begin{align}\label{lww}
l_{c}(\alpha_{j}, \Sigma_{j})={C}&+n_j \log \alpha_{j}-\frac{n_j}{2} \log |\Sigma_{j}|+\alpha_{j}\sum_{i=1}^{n_j}\log v^j_{i}-\sum_{i=1}^{n_j}
(v^j_{i})^{\alpha_{j}}\nonumber\\
&-\frac{1}{2}\sum_{i=1}^{n_j}(v^j_{i})^{2}(\boldsymbol{\cal{Y}}^{j}_{i})^T \Sigma^{-1}_j (\boldsymbol{\cal{Y}}^{j}_{i}),
\end{align}
where ${C}$ is a constant independent of $\alpha_{j}$ and $\Sigma_{j}$; for $j=1,\dots, K$.
\item{\bf{Second step of CM (first step of SEM):}} For group $I_j$, simulate the vector $\boldsymbol{V}^j=(V^{j}_{1},\dots, V^{j}_{n_j})^T$ from conditional distribution of $V^j_{i}$ given $\boldsymbol{\cal{Y}}^{j}_{i}$, $\alpha_{j}$, and $\Sigma_{j}$; for $i=1,\dots, n_j$ and $j=1,\dots, K$, using the Monte Carlo method, as described in Appendix~2, at the $N$-th cycle of stochastic step.
\item {{\bf{Third step of CM (second step of SEM):}}}
Using the vector of pseudo sample $\boldsymbol{v}^j_i=(v^{j}_{1},\dots, v^{j}_{n_j})^T$, maximize the right-hand side of (\ref{lww}) with respect to $\alpha_{j}$ and $\Sigma_{j}$ to find $\alpha_{j}^{(t+1)}$ and $\Sigma_{j}^{(t+1)}$ as
\begin{align*}\label{sigmaahat-subG}
{\Sigma}^{(t+1)}_{j}&=\frac{\sum_{i=1}^{n_j} \boldsymbol{\cal{Y}}^{j}_{i} (\boldsymbol{\cal{Y}}^{j}_{i})^T (v^j_i)^2}{n},
\end{align*}
and ${\alpha}^{(t+1)}_{j}$ is a solution of
\begin{align*}
h(\alpha_{j})=\frac{n_{j}}{\alpha_{j}}+ \sum_{i=1}^{n_{j}}\log v_i-\sum_{i=1}^{n_j} v_i^{\alpha_j} \log v_i=0,
\end{align*}
for $j=1,\dots, K$.
\par Now replace $\alpha_{j}^{(t+1)}$ and $\Sigma_{j}^{(t+1)}$ at the right-hand side of (\ref{e1ij}) and (\ref{e2ij}), respectively. This completes E-step. Then, complete M-step by updating weight and location parameters at (\ref{what}) and (\ref{muhat}). Follow three steps of the CM-steps. Repeating this loop for $N$ times, the estimated parameters of
SG$\boldsymbol{\alpha}$SM, are obtained as the following.
\begin{equation}\label{accuracy}
\hat{\lambda}_{j}=\frac{1}{N-N_0}\sum_{t=N_0+1}^{N} \lambda_{j}^{(t)},
\end{equation}
where $N_0$ is the size of burn-in for stochastic EM involved in CM-step and $\lambda_{j}^{(t)}$ is either $\alpha^{(t)}_j$, $\Sigma_{j}^{(t)}$, $\boldsymbol{\mu}^{(t)}_{j}$, or ${w}^{(t)}_j$; in $t$-th iteration of the EM algorithm for $j$-th group. Our studies reveal that setting $N=70$ and $N_0=40$ in (\ref{accuracy}) provides satisfactory accuracy in estimations.
\end{itemize}
\par
In order to implement the proposed EM algorithm, one can use the Bayesian information criterion (BIC) to estimate the number of clusters, $K$. The BIC is defined as $BIC=m\log(n)-2\log(L)$, where $\log(L)$ is the log-likelihood of observations under SG$\boldsymbol{\alpha}$SM distribution, $m$ is the number of free parameters, and $n$ is the sample size, \cite{Lee2014}. Determining $K$, to implementing the proposed EM algorithm, we first use package \textsf{cluster}, \cite{Maechler2015}, for pre-clustering and then package $\mathsf{STABLE}$, \cite{stable}, for initial estimates of the parameters within each cluster.
\section{Simulation study and real data analysis}
\begin{figure}[h]
\centering
\includegraphics[width=7cm,height=7cm]{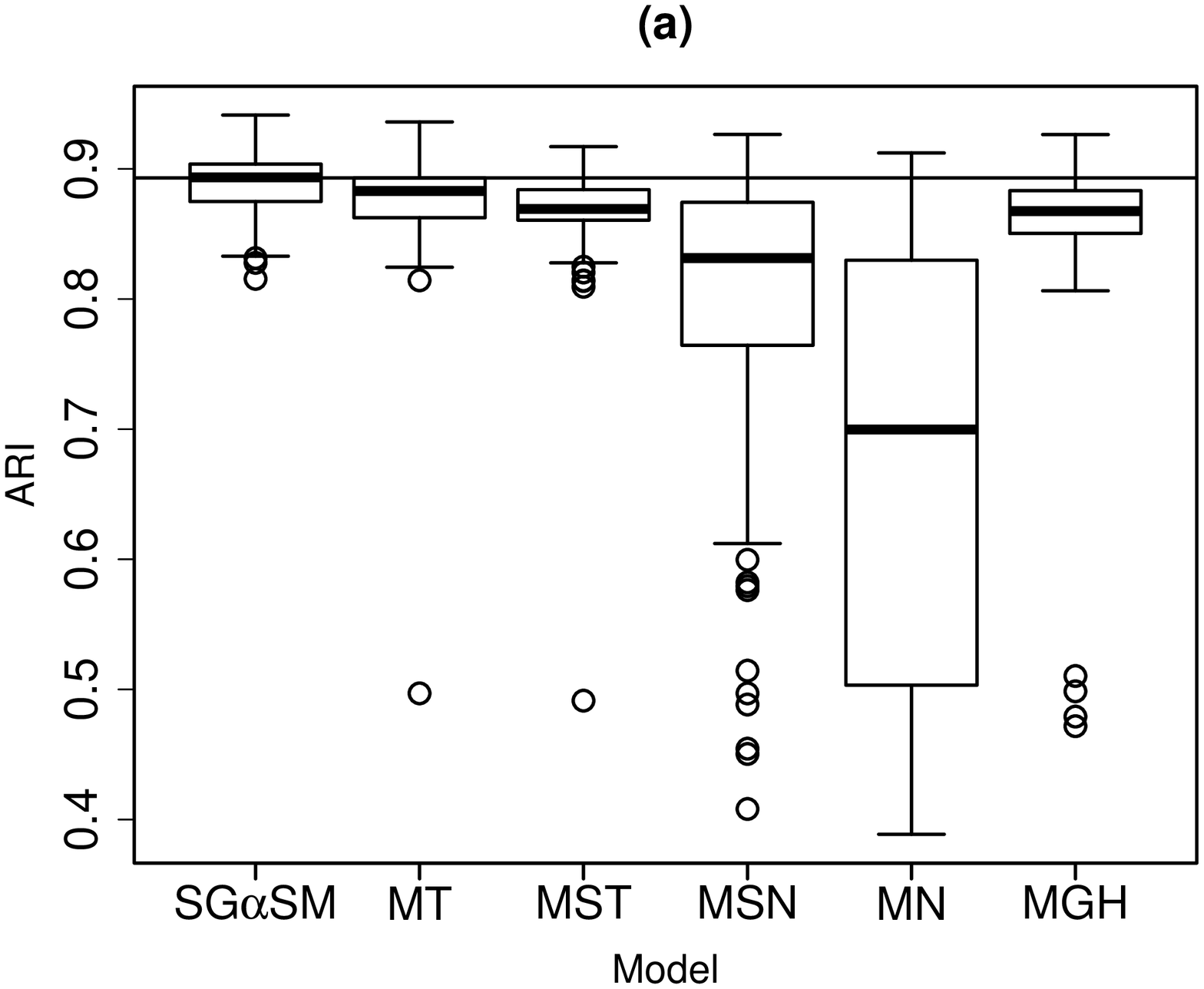}
\includegraphics[width=7cm,height=7cm]{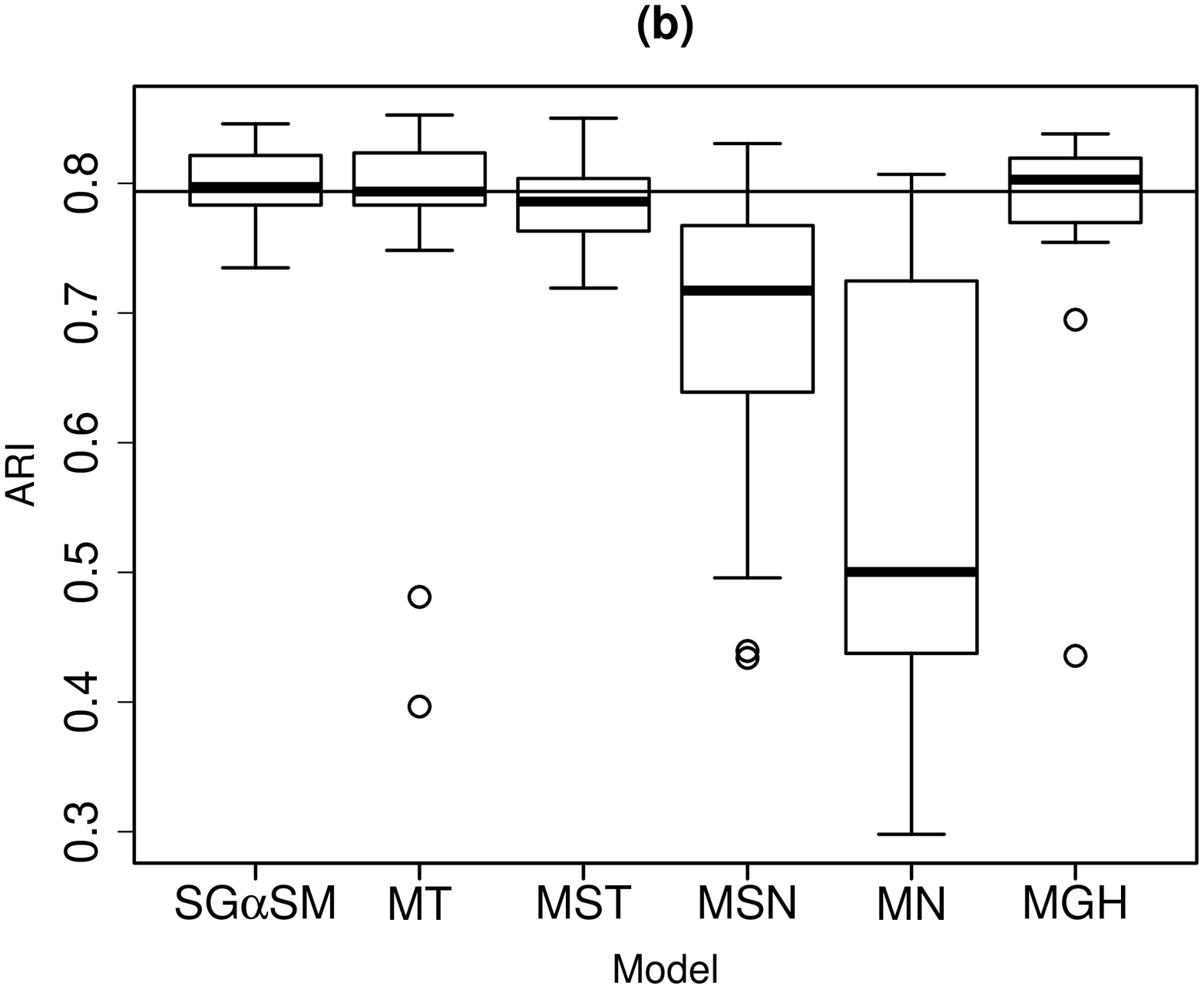}
\caption{Computed ARI when: (a) data generated from 3-component SG$\boldsymbol{\alpha}$SM model and (b) data generated from 3-component MT. The horizontal lines in left- and right-hand sides show the median of ARI when data are modelled with SG$\boldsymbol{\alpha}$SM distribution and MT, respectively.}\label{Fig1}
\end{figure}

\begin{figure}[h]
\centering
\includegraphics[width=7cm,height=7cm]{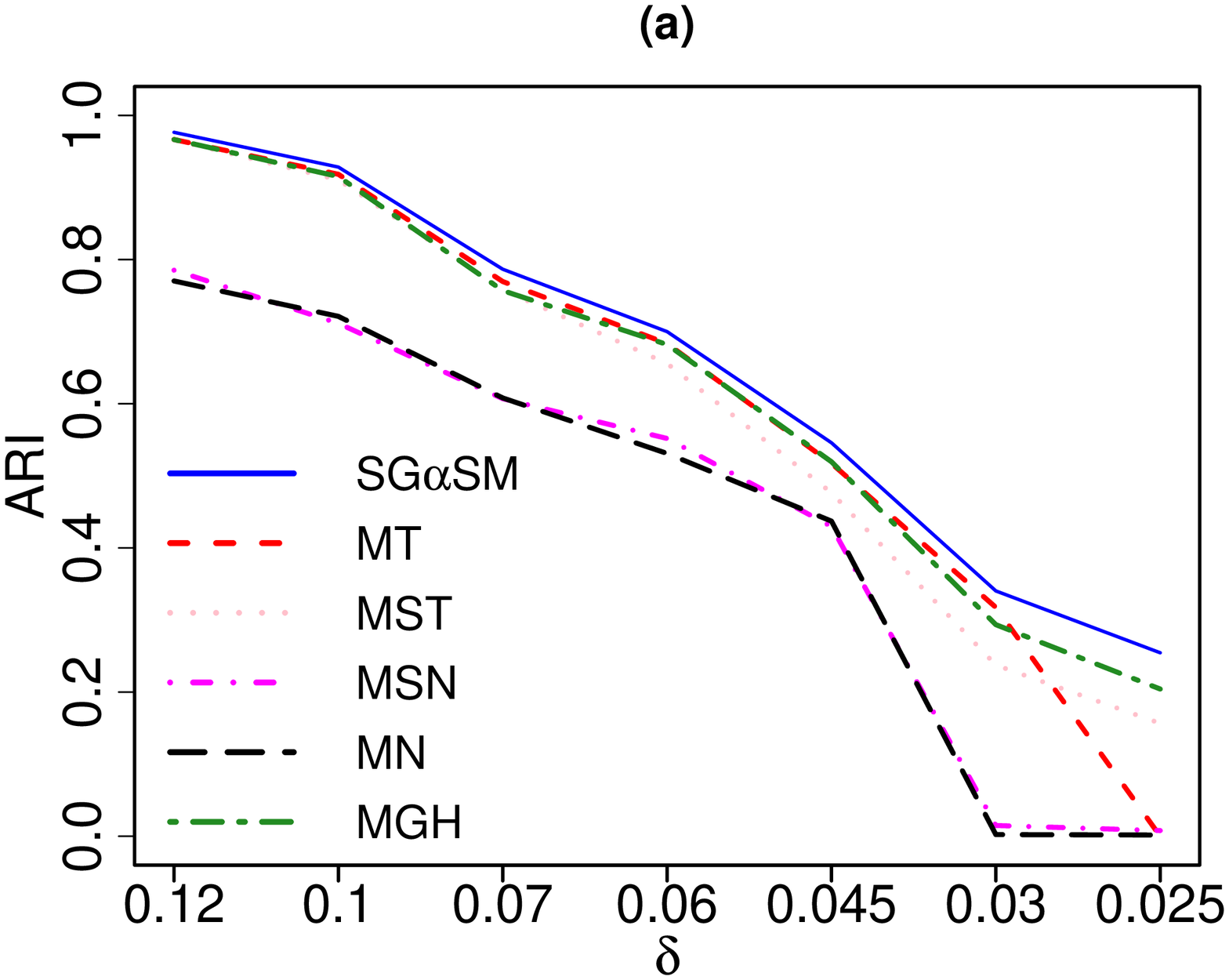}
\includegraphics[width=7cm,height=7cm]{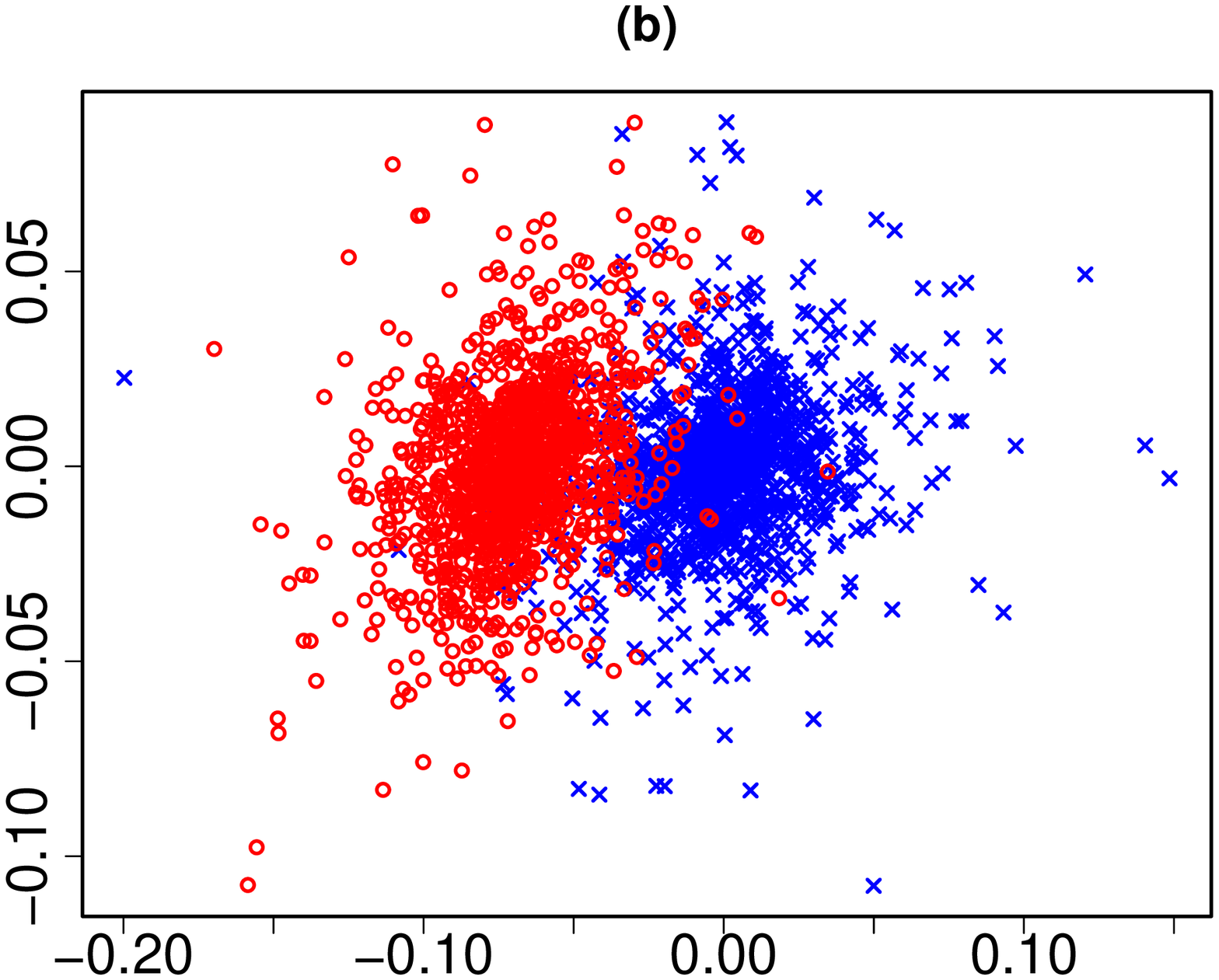}
\caption{Left panel (a): computed ARI when 2-component SG$\boldsymbol{\alpha}$SM model, MT, MST, MN, MSN, and MGH are fitted to the 2494 mixed samples of $\boldsymbol{X}_1=(JPM,MCD)^T$ and $\boldsymbol{X}_2=(AXP-\delta,SBC)^T$ for different values of $\delta$. Right panel (b): scatter plot of mixed observations for $\delta=0.07$. Red circle- and blue star-shaped points correspond to $\boldsymbol{X}_2$ and $\boldsymbol{X}_1$, respectively.}\label{Fig2}
\end{figure}

This section has two parts. Firstly, we compare the performance of the proposed EM algorithm to modelling data simulated from SG$\boldsymbol{\alpha}$SM distribution. We also check the robustness of SG$\boldsymbol{\alpha}$SM distribution when data generated from a mixture of $t$ (MT) distribution in the presence of a mixture of skew $t$ (MST), a mixture of normal (MN), a mixture of skew normal (MSN), and a mixture of generalized hyperbolic (MGH) distributions. Secondly, for synthetic data analysis, we choose four stocks among 30 stocks of Dow Jones data, \cite{Nolan2013}. Finally, for real data analysis, we focus on $\mathsf{banknote}$ data set which involves of six variables made on 100 genuine and 100 counterfeit Swiss bank notes. This dataset is available by loading package $\mathsf{MixGHD}$, \cite{Tortora2015}. It should be noted that, during analyses, we use package $\mathsf{mixsmsn}$ to model data via MT, MST, MN, and MSN distributions, \cite{Prates2013}. Also, package $\mathsf{MixGHD}$ is applied for modelling data by MGH distribution and computing adjusted Rand index (ARI) as a measure of performance, \cite{Hubert1985}.
\\\\{\bf{Example 1:}} Performance of the SG$\boldsymbol{\alpha}$SM distribution is investigated through a small simulation study. For performance, we simulate 200 times realizations from 3-component SG$\boldsymbol{\alpha}$SM distribution under settings: $n=600$, $\boldsymbol{w}=(1/3,1/3,1/3)^T$, $\boldsymbol{\mu}_1=(0,3)^T$, $\boldsymbol{\mu}_2=(3,0)^T$, $\boldsymbol{\mu}_3=(-3,0)^T$,
$\boldsymbol{\Sigma}_1=\begin{pmatrix}2&0.5\\ 0.5&0.5\\ \end{pmatrix}$, $\boldsymbol{\Sigma}_2=\begin{pmatrix}1&0\\ 0&1\\ \end{pmatrix}$, and $\boldsymbol{\Sigma}_3=\begin{pmatrix}2&-0.5\\ -0.5&0.5\\ \end{pmatrix}$. We choose these settings of parameters from \cite{Peel2000}. In following, Figure \ref{Fig1} displays the ARI computed under each of six mixture models. AS it is expected, Figure \ref{Fig1}(a), the SG$\boldsymbol{\alpha}$SM model shows the best performance in the sense of median of ARI. In order to investigate the robustness of the SG$\boldsymbol{\alpha}$SM model, we simulate 200 times realizations from 3-component MT distribution under above settings (the degrees of freedoms for first, second, and third components are $\nu_1=2$, $\nu_2=4$, and $\nu_3=8$, respectively). Figure \ref{Fig1}(b) shows the computed ARIs when data are generated from MT distribution. As it is seen, surprisingly, MGH and SG$\boldsymbol{\alpha}$SM model shows better performance than MT based on the median of ARI.
\\\\{\bf{Example 2:}} We choose stocks AXP, JPM, MCD, and SBC stocks from Dow Jones data. It can be checked that a symmetric $\alpha$-stable pdf captures well the empirical distribution of these stocks based on 1247 observations with almost the same tail indices. Also, the joint scatterplots of these stocks are roughly elliptical. This means that a SG$\alpha$S distribution is suitable for modelling these stocks, \cite{Nolan2013}. Define $\boldsymbol{X}_1=(\text{JPM},\text{MCD})^T$ and $\boldsymbol{X}_2=(\text{AXP}-\delta,\text{SBC})^T$. The scatterplots of $\boldsymbol{X}_1$ and $\boldsymbol{X}_2$ have a perfect overlay when $\delta$ is zero and are well-separated when $\delta$ is large (say $\delta>0.3$). In the following, Figure \ref{Fig2} displays the computed ARI for $\delta$=0.12,0.1,0.07,0.06,0.045,0.03,0.025. As it is seen, SG$\boldsymbol{\alpha}$SM distribution shows the best performance.
\\\\{\bf{Example 3:}} Among variables, we choose the width of left edge (left) and bottom margin width (bottom) from $\mathsf{banknote}$ data, \cite{Fraley2016}. Computed ARIs correspond to SG$\boldsymbol{\alpha}$SM model, MT, MST, MN, MSN, and MGH are 0.721102, 0.704122, 0, 0.704122, 0, and 0.7041185, respectively. This report indicates that SG$\boldsymbol{\alpha}$SM gives the best performance.
\section{Conclusion remarks}
The E-step of the EM algorithm for calculating maximum likelihood estimates of the sub-Gaussian $\alpha$-stable mixture (SG$\boldsymbol{\alpha}$SM) distribution parameters is not tractable computationally. We propose here some methodology that makes it possible to evaluate the intractable E-step for the SG$\boldsymbol{\alpha}$SM distribution. We assume that the number of components is known and starting values for the EM algorithm are estimated using statistical packages have provided for clustering. A simulation study reveals that the proposed EM algorithm is robust against to starting values, outliers, and deviations from model assumptions. This is proved when data are generated from a mixture of $t$ distributions. Also, the performance of the proposed EM algorithm is demonstrated using synthetic and real data. We hope practitioners find this model useful for practical purposes. As a possible future work, we would like to develop the methodology proposed here to operator SG$\boldsymbol{\alpha}$SM distribution in which, components of each cluster have different tail weights.
\section*{Appendix 1}
At $(t+1)$-th iteration of the E-step, to compute $e^{(t+1)}_{1ij}$, we need computing the pdf of a SG$\alpha$S, i.e. $f\bigl({\boldsymbol{y}}_{i};\alpha^{(t)}_j, \Sigma^{(t)}_j, \boldsymbol{\mu}^{(t)}_{j}\bigr)$. Also, it can be checked that
\begin{align*}
e^{(t)}_{2ij}=\frac{\int_{0}^{\infty}u^{-d/2-1}f_P\bigl(u|\alpha^{(t)}_j\bigr)\exp\left(-\frac{\bigl(\boldsymbol{y}_i-\boldsymbol{\mu}^{(t)}_j\bigr)^T \bigl(\Sigma^{(t)}_j\bigr)^{-1} \bigl(\boldsymbol{y}_i-\boldsymbol{\mu}^{(t)}_j\bigr)}{2u}\right)du}{\int_{0}^{\infty}u^{-d/2}f_P\bigl(u|\alpha^{(t)}_j\bigr)\exp\left(-\frac{\bigl(\boldsymbol{y}_i-\boldsymbol{\mu}^{(t)}_j\bigr)^T \bigl(\Sigma^{(t)}_j\bigr)^{-1} \bigl(\boldsymbol{y}_i-\boldsymbol{\mu}^{(t)}_j\bigr)}{2u}\right)du}.
\end{align*}
Both of $e^{(t+1)}_{1ij}$ and $e^{(t+1)}_{2ij}$ are computed using package $\mathsf{STABLE}$.
\section*{Appendix 2}
To simulate the pseudo-complete data from conditional distribution of $V^j_{i}$ given $\boldsymbol{\cal{Y}}^{j}_{i}$, $\alpha_{j}$, and $\Sigma_{j}$; for $i=1,\dots, n_j$ and $j=1,\dots, K$, we use rejection sampling by the following steps. We have our idea from \cite{Gonzalez2010} as follows. We note that the density function
\begin{equation*}
f_{\boldsymbol{\cal{Y}}^{j}_{i}|V^j_{i},\alpha_{j},\Sigma_{j}}(\boldsymbol{\cal{Y}}^{j}_{i}|v^j_{i},\alpha_{j},\Sigma_{j})=\frac{\bigl(v^j_{i}\bigr)^{d}}{(2\pi)^{d/2}|\Sigma_{j}^{(t)}|^{1/2}} \exp\left\{-\frac{\bigl((\boldsymbol{\cal{Y}}^{j}_{i})^T
\Sigma^{-1}_j\boldsymbol{\cal{Y}}^{j}_{i}\bigr)\bigl(v^j_{i}\bigr)^2}{2}\right\},
\end{equation*}
as a part of conditional (posterior) density function
\begin{align*}
f_{V^j_{i}|\boldsymbol{\cal{Y}}^{j}_{i},\alpha_{j},\Sigma_{j}}(v^j_{i}|\boldsymbol{\cal{Y}}^{j}_{i},\alpha_{j},\Sigma_{j})
&\propto f_{V^j_{i}}(v^j_{i}) f_{\boldsymbol{\cal{Y}}^{j}_{i}|V^j_{i},\alpha_{j},\Sigma_{j}}(\boldsymbol{\cal{Y}}^{j}_{i}|v^j_{i},\alpha_{j},\Sigma_{j}),\nonumber\\
&=\frac{\alpha\bigl(v^j_{i}\bigr)^{d+\alpha -1}}{(2\pi)^{d/2}|\Sigma_{j}^{(t)}|^{1/2}} \exp\left\{-\frac{\bigl((\boldsymbol{\cal{Y}}^{j}_{i})^T
\Sigma^{-1}_j\boldsymbol{\cal{Y}}^{j}_{i}\bigr)\bigl(v^j_{i}\bigr)^2}{2}-\bigl(v^j_{i}\bigr)^{\alpha}\right\},
\end{align*}
is bounded by some constant independent of $v^j_{i}$. More precisely, by differentiating density $f_{\boldsymbol{\cal{Y}}^{j}_{i}|V^j_{i},\alpha_{j},\Sigma_{j}}(\boldsymbol{\cal{Y}}^{j}_{i}|v^j_{i},\alpha_{j},\Sigma_{j})$ with respect to $v^j_{i}$, it turns out that
$f_{\boldsymbol{\cal{Y}}^{j}_{i}|V^j_{i},\alpha_{j},\Sigma_{j}}(\boldsymbol{\cal{Y}}^{j}_{i}|v^j_{i},\alpha_{j},\Sigma_{j})$ attains its maximum as
\begin{align*}
\frac{\exp\left\{-\frac{d}{2}\right\}\bigl(\frac{d}{(\boldsymbol{\cal{Y}}^{j}_{i})^T
\Sigma^{-1}_j\boldsymbol{\cal{Y}}^{j}_{i}}\bigr)^{\frac{d}{2}}}{(2\pi)^{d/2}|\Sigma_{j}^{(t)}|^{1/2}},
\end{align*}
at point
$v^j_{i}=\sqrt{\frac{d}{((\boldsymbol{\cal{Y}}^{j}_{i})^T
\Sigma^{-1}_j\boldsymbol{\cal{Y}}^{j}_{i})}}$. Hence, the rejection sampling approach is employed to generate from the posterior distribution by the following steps.
\begin{enumerate}
\item Simulate a sample, say $v^j_{i}$, from a Weibull distribution with shape parameter $\alpha_j$ and scale unity.
\item Define $b=\frac{d^{d/2}\left((\boldsymbol{\cal{Y}}^{j}_{i})^T
\Sigma^{-1}_j\boldsymbol{\cal{Y}}^{j}_{i}\right)^{-d/2} \exp\{-d/2\}}{(2\pi)^{d/2}| \Sigma_j |^{1/2}}$ and generate a sample from a uniform distribution $U\left(0,b\right)$, say $u$.
\item If $u<\frac{(v^j_{i})^{d}\exp\{-\frac{1}{2}\left((\boldsymbol{\cal{Y}}^{j}_{i})^T
\Sigma^{-1}_j\boldsymbol{\cal{Y}}^{j}_{i}\right)(v^j_{i})^2\}}{(2\pi)^{d/2}| \Sigma_j |^{1/2}}$,
then accept $v^j_{i}$ as an observation pdf $V^j_{i}$ given $\boldsymbol{\cal{Y}}^{j}_{i}$, $\alpha_{j}$, and $\Sigma_{j}$; for $i=1,\dots, n_j$ and $j=1,\dots, K$; otherwise, go to step 1.
\end{enumerate}


\begin{thebibliography}{99}
\bibitem{Andrews2012}
Andrews, J. L. and McNicholas, P. D. (2012). Model-based clustering, classification, and discriminant analysis via mixtures of multivariate $t$-distributions. \emph{Statistics and Computing}, 22, 1021-1029.
\bibitem{Basso2010}
Basso, R. M., Lachos, V. H. Cabral, C. R. B., and Ghosh, P. (2010). Robust mixture modeling based on scale mixtures of skew-normal distributions, \emph{Computational Statistics and Data Analysis}, 54, 2926-2941.
\bibitem{Bodnar2011}
Bodnar, T. and Gupta, A. K. (2011). Estimation of the precision matrix of a multivariate elliptically contoured stable distribution,
\emph{Statistics}, 45(2), 131-142.
\bibitem{Browne2015}
Browne, R. P. and McNicholas, P. D. (2015). A mixture of generalized hyperbolic distributions. \emph{The Canadian Journal of Statistics}, 43(2), 176-198.
\bibitem{Cabral2012}
Cabral, C. R. B., Lachos, V. H., and Prates, M. O. (2012). Multivariate Mixture Modeling Using
Skew-Normal Independent Distributions. \emph{Computational Statistics and Data Analysis}, 56, 126-142.
\bibitem{Celeux1985}
Celeux, G. and Diebolt, J. (1985). The SEM algorithm: a probabilistic teacher algorithm derived from the EM algorithm for mixture problem, \emph{Computational Statistics Quarterly}, 2 (1), 73-82.
\bibitem{Dempster1977}
Dempster, A. P., Laird, N. M., and Rubin, D. B. (1977). Maximum likelihood from incomplete data via the EM algorithm, \emph{Journal of the Royal Statistical Society, Series B}, 39, 1-38.
\bibitem{Fraley2016}
Fraley, C., Raftery, A. E., and Scrucca, L. (2016). {\it{\textsf{mclust}}}: Normal Mixture Modeling for Model-Based Clustering, ?Classification, and Density Estimation, R package version 5.2.
\bibitem{Holzmann2006}
Holzmann, H., Munk, A. and Gneiting, T. (2006). Identifiability of finite mixtures of elliptical distributions, \emph{Scandinavian Journal of Statistics}, 33, 753-763.
\bibitem{Hubert1985}
Hubert, L. J. and Arabie, P. (1985). Comparing partitions, \emph{Journal of Classification}, 2, 193-218.
\bibitem{Ip1994}
Ip, E. H. S. (1994). A stochastic EM estimator for handling missing data, Unpublished Ph.D. Thesis, Department of Statistics, Stanford University.
\bibitem{Kring2008}
Kring, S., S. T. Rachev, M. H\"{o}chst\"{o}tter, and F. J. Fabozzi (2009). \emph{Risk Assessment: Decisions in Banking and Finance}, Chapter 6: Estimation of $\alpha$-stable sub-Gaussian distributions for asset returns, 111-152, Springer/Physika, Heidelberg.
\bibitem{Lee2013}
Lee, S. and McLachlan, G. J. (2013). On mixtures of skew normal and skew $t$-distributions, \emph{Advances in Data Analysis and Classification}, 7, 241-266.
\bibitem{Lee2014}
Lee, S. and McLachlan, G. J. (2014). Finite mixtures of multivariate skew $t$-distributions: some recent and new results, \emph{Statistics and Computing}, 24, 181-202.
\bibitem{Liu1994}
Liu, C. and Rubin, D. B. (1994). The ECME algorithm: A simple extension of EM and ECM with faster monotone convergence, \emph{Biometrika}, 81, 633-648.
\bibitem{Maechler2015}
Maechler, M., Rousseeuw, P., Struyf, A., Hubert, M., and Hornik, K. (2015). {\it{\textsf{cluster}}}: Cluster Analysis Basics and Extensions, R package version 2.0.1.
\bibitem{Meng1993}
Meng, X. L. and Rubin, D. B. (1993). Maximum likelihood estimation via the ECM algorithm: A general framework, \emph{Biometrika}, 80, 267-278.
\bibitem{Mittnik2003}
Mittnik, S. and Paolella, M. S. (2003). Prediction of financial downside risk with heavy tailed conditional distributions, In Rachev, S. T. (Ed.), Handbook of Heavy Tailed Distributions in Finance, Elsevier Science, Amsterdam.
\bibitem{Nikias1995}
Nikias, C. L. and Shao, M. (1995). \emph{Signal Processing with $\alpha$-Stable Distributions and Applications}, John Wiley, New York.
\bibitem{Nolan2003}
Nolan, J. P. (2003). \emph{Modeling financial distributions with stable distributions}, Volume 1 of \emph{Handbooks in Finance},
Chapter 3, pp. 105-130, Elsevier, Amsterdam.
\bibitem{Nolan2013}
Nolan, J. P. (2013). Multivariate elliptically contoured stable distributions: theory and estimation, \emph{Computational Statistics}, 28, 2067-2089.
\bibitem{Prates2013}
Prates, M. O., Cabral, C. R. B., and Lachos, V. H. (2013). {\it{\textsf{mixsmsn}}}: Fitting
Finite Mixture of Scale Mixture of Skew-Normal Distributions. \emph{Journal of Statistical Software}, 54(12), 1-20.
\bibitem{Peel2000}
Peel, D. and McLachlan, G. J. (2000). Robust mixture modelling using the $t$ distribution, \emph{Statistics and Computing}, 10, 339-348.
\bibitem{Ortobelli2010}
Ortobelli, S., Rachev, S. T., and Fabozzi, F. J. (2010). Risk management and dynamic portfolio selection with stable Paretian distributions, \emph{Journal of Empirical Finance}, 17(2), 195-211.
\bibitem{Omelchenko2010}
Omelchenko, V. (2010). Elliptical stable distributions, In Houda, M. and Friebelova, J. (Eds.), \emph{Mathematical Methods in Economics}, 483-488.
\bibitem{Omelchenko2014}
Omelchenko, V. (2014). Parameter estimation of sub-Gaussian stable distributions, \emph{Kybernetika}, 50(6), page 929-949.
\bibitem{Rachev2000}
Rachev, S. T. and Mittnik, S. (2000). \emph{Stable Paretian Models in Finance}, John Wiley, New York.
\bibitem{Ravishankera1999}
Ravishankera, N. and , Qiou, Z. (1999). Monte Carlo EM estimation for multivariate stable distributions, \emph{Statistics and Probability Letters}, 45(4), 335-340.
\bibitem{stable}
Robust Analysis Inc (2010). \emph{User Manual for STABLE 5.0}. Software and user manual available online at www.RobustAnalysis.com
\bibitem{Gonzalez2010}
Salas-Gonzalez, D., Kuruoglu, E. E., and Ruiz, D. P. (2010). Modelling with mixture of symmetric stable distributions using Gibbs sampling, \emph{Signal Processing}, 90, 774-783.
\bibitem{Samorodnitsky1994}
Samorodnitsky, G. and Taqqu, M. S. (1994). \emph{Stable Non-Gaussian Random Processes: Stochastic Models and Infinite Variance}, Chapman and Hall, London.
\bibitem{Tortora2015}
Tortora, C., Browne, R. P., Franczak, B. C., and McNicholas, P. D. (2015). {\it{\textsf{MixGHD}}}: Model based clustering, classification and discriminant analysis using the mixture of generalized hyperbolic distributions, R package version 1.8.
\bibitem{Vrbik2012}
Vrbik, I. and McNicholas, P. D. (2012). Analytic calculations for the EM algorithm for multivariate skew-$t$ mixture models, \emph{Statistics and Probability Letters}, 82, 1169-1174
\bibitem{Zhang2010}
Zhang, J. and Liang, F. (2010). Robust clustering using exponential power mixtures, \emph{Biometrics}, 66, 1078-1086.
\end{thebibliography}
\end{document}